# Rethinking Trust Repair in Human-Robot Interaction

**Connor Esterwood**
School of Information
University of Michigan
Ann Arbor, MI 48104
`cte@umich.edu`

July 14, 2023

## Abstract

As robots become increasingly prevalent in work-oriented collaborations, trust has emerged as a critical factor in their acceptance and effectiveness. However, trust is dynamic and can erode when mistakes are made. Despite emerging research on trust repair in human-robot interaction, significant questions remain about identifying reliable approaches to restoring trust in robots after trust violations occur. To address this problem, my research aims to identify effective strategies for designing robots capable of trust repair in human-robot interaction (HRI) and to explore the underlying mechanisms that make these strategies successful. This paper provides an overview of the fundamental concepts and key components of the trust repair process in HRI, as well as a summary of my current published work in this area. Additionally, I discuss the research questions that will guide my future work and the potential contributions that this research could make to the field.

## 1 Introduction

Robots are sophisticated machines that combine artificial intelligence with the materiality of a physical body. This unique combination of artificial mind and body often leads to the positioning of robots not only as tools but also as teammates and collaborators [1, 2]. As a result, robots have increasingly been deployed in contexts historically reserved for humans often as vital members of work-oriented teams [3, 4, 5, 6]. However, for these teams to be effective, the establishment and maintenance of trust is vital [7, 8, 9, 10, 11]. This is because trust is the "glue that holds most cooperative relationships together" [12, Pg.129].

Trust can be formally defined as the willingness of the trustor to be vulnerable to the actions of the trustee [13]. To date, a wealth of literature exists on human–robot trust and extensive reviews and meta-analyses have showcased how we as a community have come to understand what leads humans to – or indeed not to – trust robots [14, 15]. This produces a strong base of literature that can be leveraged but, the dynamic nature of trust is often overlooked.

Trust is rarely stable but instead changes over time based on past and current interactions [16, 17, 18, 19]. Increases in trust are relatively easy to manage, decreases, however, can have lasting effects and are significantly harder to recover from [19]. Fortunately, there are numerous strategies that humans deploy in order to restore trust and therefore act as useful tools for preserving relationships and ensuring the continued efficacy of teams and collaborative work arrangements.

Trust repair strategies can be defined as planned approaches or sets of actions designed to rebuild trust after it has been damaged or violated [20, 21]. While researchers in human-robot interaction (HRI) have started investigating the effectiveness of these strategies when employed by robots, a recent literature review revealed mixed results regarding their efficacy [22]. Moreover, a comprehensive theoretical framework for trust repair in HRI is yet to be established, and the specific mechanisms underlying different repair techniques remain largely unexplored. Consequently, there is still a lack of knowledge necessary to empower designers and developers to construct robots capable of predictably and reliably restoring trust following trust violations.

Recent studies have examined the efficacy of trust repair strategies when deployed by robots. Findings, however, have been mixed, and there is a lack of a comprehensive theoretical framework for trust repair in HRI [22]. As a result, designers and developers still lack the knowledge needed to build robots that can effectively and consistently restore trust after violations occur. To address this gap, my current and future research works towards the broader goal of developing a theoretical model of trust repair. In doing so it explores how trust repairs function and why some repairs are effective at certain times and others are not. The results of this research empower designers to more intelligently design robots with the capacity for trust repair and as a result encourage the kinds of collaborative relationships necessary for this novel technology to be accepted and useful.

This extended abstract is structured as follows. First, it will introduce the concepts of trust and trustworthiness. Then, it will discuss trust repair and summarize existing findings from the HRI trust repair literature. Next, it will present the methodologies I have developed to explore trust repair in HRI. Finally, it will provide a summary of my existing work on this subject, as well as my future planned work, which will culminate in my dissertation.

## 2 Background

Trust, which is vital to collaboration [23], is largely preceded by trustworthiness [13, 24]. Specifically, trustors (agents bestowing trust) evaluate the trustworthiness of the trustee (agent receiving trust) over time, learning from the outcomes of their interactions. In this sense, trustworthiness can be viewed as learned rather than dispositional or situational. Trustworthiness has been shown to play a significant role in shaping human's trust in robots [25].

Trustworthiness can be subdivided into three different components, namely, ability, integrity, and benevolence. Ability refers to the extent to which the trustee is perceived as skilled or competent in a specific domain [13]. Integrity relates to the degree to which the trustee is viewed as honest and adherent to a set of principles [25]. Lastly, benevolence pertains to the degree to which the trustee is regarded as acting selflessly without any conflicting egocentric or profit-based motives [13]. Each of these different sub-components can have unique impacts on how a trustor perceives a trustee and as a result influences the willingness of a trustor to be vulnerable to said trustee. Importantly, these perceptions are malleable allowing a trustee to influence how they are seen by a trustor. One way that trustees often manipulate these perceptions is through the deployment of different trust repair strategies.

### 2.1 Trust Repairs

Trust repairs are actions taken by a trustee to restore trust after a perceived or actual violation of trust [26, 27, 20]. These actions can take various forms but are often realized through short-term verbal repair strategies, such as apologies, denials, explanations, and promises [19, 28, 22]. Each of these trust repair strategies can be linked to different dimensions of trustworthiness and are supported by distinct overarching theoretical frameworks [13, 24, 21].

*Apologies* are expressions of regret or remorse for a perceived transgression. They are the most widely used trust repair strategy across the literature and are believed to repair trust through encouraging forgiveness [21, 22]. *Denials* are attempts to redirect blame or reject culpability for a trust violation [29], aiming to establish the complete innocence of the trustee by shifting blame onto another entity [19]. Denials are hypothesized to work through misinforming because they rely largely on inaccurate information provided by the trustee to the trustor [21].

*Explanations*, on the other hand, provide clear reasoning behind a trust violation, seeking to establish a shared understanding between the trustor and trustee by conveying transparency [19, 21, 30, 31]. Explanations are hypothesized to repair trust through informing the trustor [21]. Finally, *promises* are statements of commitment to positive future performance [32], aiming to restore trust by shifting the focus from past to future behaviors and believed to work through encouraging forgetfulness [21].

In the HRI literature, the effectiveness of trust repair strategies has produced mixed results [22]. It is therefore possible that various moderating factors, such as timing [33, 34], the number of repeated trust violations [35], violation type [36, 37], anthropomorphism [38], attitude [35], and violation severity [39], can influence the efficacy of these strategies. The field of trust repair in HRI, however, is still in its early stages, and there is much that remains to be explored. Additional moderators or previously unexamined main effects may more clearly explain the divergent results in the literature. Additionally, human-related factors and individual differences may play a significant role in determining the efficacy of trust repairs. Moreover, the theoretical links between trustworthiness and individual trust repair strategies have yet to be fully established and explored.

Leveraging this existing work and keeping the theoretical links between trustworthiness, trust, and trust repair in mind I have positioned my research to examine what factors make repairs more or less effective, when this is the case, and why. I have done so by first conducting a series of studies examining trustworthiness, positive attitudes, and anthropomorphism. These studies results are then leveraged to inform my forthcoming work on trust repair related to personality congruence and mind perception as well as my future planned work that seeks to examine the sub-components of trustworthiness in greater detail and how different combinations and the ordering of apologies, denials, explanations, and promises might lead to more or less effective trust repairs. I now transition to a brief introduction of the methodologies used for this research followed by a summary of prior work, my planned dissertation work, and the potential contributions of these investigations.

## 3 Methodological Development & The Warehouse Robot Interaction Sim

To examine if, when, and how robots repair trust, I first needed to create a scenario where robots violate it. I achieved this by designing and developing an open-source platform called The Warehouse Robot Interaction Sim (WRIS) using the Unreal Engine 4.27. The WRIS consists of a cooperative human-robot interaction task where one human teamed up with one robot in an immersive virtual environment. This environment is visible in figure 1 and contained a table, two monitors, a stack of boxes, a conveyor belt, and a robot.

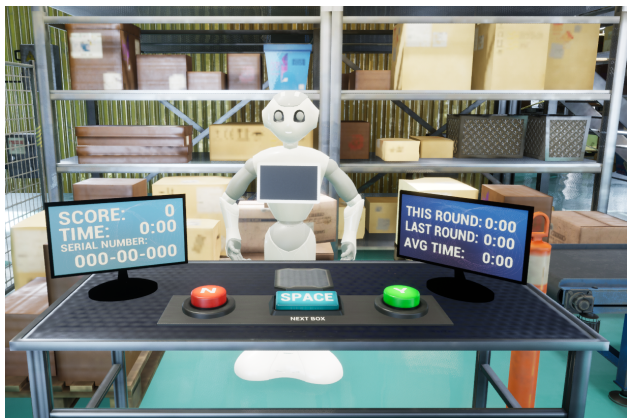

Figure 1: The Warehouse Robot Interaction Sim as seen by participants.

Within this environment participants are assigned as a member of a one human to one robot team. The team's objective is to correctly process and place ten boxes on a nearby conveyor belt. The boxes are processed correctly if the serial number on the box matches the serial number on a nearby display. It is the human's responsibility to determine whether the boxes were correct or incorrect, and it is the robot's responsibility to pick a matching box, present it to the human, and move it to the appropriate location based on the human's feedback. When the team processes a correct box their score is incremented by 1 point. When the team processes an incorrect boxes their score is decremented by 1 point.

The team is timed, and the time taken to process each box and the average time taken to process boxes is also visible to participants. Trust violations in the WRIS take the form of a robot selecting an incorrect box to present to participants. This increases their total time on task and reduces the total number of points they can score during the simulation. By making both timing and scores visible to participants the WRIS encourages participant engagement and make the robot's presentation of incorrect boxes (i.e. trust violations) more consequential. Figure 2 illustrates this task and the different choices available to humans and their associated consequences.

Alongside the WRIS I deployed a collection of different measures, the most important of these measures being a 3-item trust measure modified from [40] and a 9-item trustworthiness measure based on [41, 42, 43]. These measures allowed me to first examine how trust changed over multiple violations and second to assess how trustworthiness and its sub-components were impacted after the study concluded. These items are detailed in associated publications (See: [21, 38, 35]) and were found to be sufficiently reliable. However, future work will seek to iterate upon these questionnaires and further improve their reliability.

## 4 Prior Work

Using the above methodology, I have gained valuable insights into trust repair in HRI through my research. Specifically, a recent series of studies [38, 21, 35] have shown that the effectiveness of different trust repair strategies can vary depending on robot type (i.e. anthropomorphism), repair timing, and individual differences in positive attitudes towards robots.

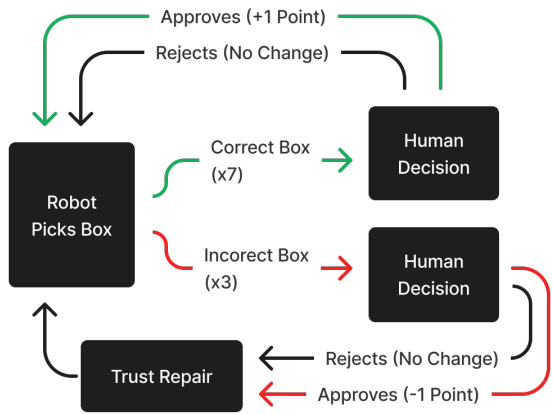

Figure 2: A visual representation of the task flow used in The Warehouse Robot Interaction Sim.

The first study in this series [38] examined the impact of anthropomorphism on trust repair. This study aimed to determine if apologies, denials, explanations, and promises had unique impacts on trustworthiness when given by a machine-like robot as opposed to a human-like robot. The results of this study revealed that a robot's human-likeness (i.e. anthropomorphism) impacts certain aspects of the trust repair process, but this impact is likely variable and potentially nonlinear. For example, apologies, denials, and promises given by the human-like robot resulted in lower perceptions of integrity than when given by the machine-like robot. Similarly, human-like robots offering promises were seen as less benevolent than machine-like robots offering the same repair.

Furthermore, [21] examined how robust trust repair strategies are against repeated trust violations. Results indicated that after 3 trust violations none of the repair strategies examined ever fully repaired perceptions of a robot's ability or integrity. In addition, after repeated interactions, apologies, explanations, and promises appeared to function similarly to one another while denials were consistently the least effective. One limitation of both of these earlier studies, however, is that trustworthiness was only measured once. This prevented the examination of the impact of repeated trust violations in a more direct manner.

To date, no study has examined trustworthiness and it's sub-components in light of repeated violations of trust. However, previous work has examined how repeated violations impact trust and how positive attitude may moderate these impacts. In particular, [35] examined the impact of apologies, denials, explanations, and promises after a 1st, 2nd, and 3rd violation of trust occurred. This produced three important findings. First, This study found that positive attitudes do appear to moderate the impact of different trust repair strategies. Second, it also observed that this impact varied between strategies and was only significant for promises. Finally, and most important given the gap above, this study also found that trust was only significantly impacted by repairs after the 2nd violation. This provides some early evidence supporting the idea that repeated violations may have an impact on the efficacy of different trust repairs but, it does not explain how nor does it elucidate what mechanisms of trust repair may be impacted and when.

## 5 Planned Work

Overall my previous work on trust repair in HRI provides novel insights but also raise several new questions. As a result, I have built upon this previous work and developed a series of three important research questions to be incorporated into my dissertation research. Like my previous work in this area, these studies will leverage the WRIS and will examine trust violations and repairs delivered by the robots within it. However, unlike my previous work, these studies will examine characteristics of the trust repair strategies themselves and the potential mechanisms through which they function. This represents a departure from my previous work, which primarily focused on individual differences. I will now briefly introduce each of these planned studies, their associated research questions, and potential contributions to the literature.

### 5.1 Study 1 - Mechanisms, Trustworthiness and Repeated Trust Violations

In my previous examinations of trust repair in HRI I found that different trust repair strategies appear to have unique impacts on trustworthiness. Within this work, however, I did not examine how these unique impacts may be influenced by repeated violations beyond the overall level (i.e. after the 1st, 2nd, and 3rd violations of trust respectively). This presents an important gap in the literature as we do not yet know if these trust repairs were effective after earlier violations occurred and if the mechanisms proposed by [21] manifest only after some violations but not others. This leads me to my first research question:

> **Study 1 | RQ:** Do specific repair strategies restore trust by recovering perceptions of a robot's ability, integrity and/or benevolence? If so, do these impacts become more of less effective over the course of repeated trust violations?

This question is important to answer for two reasons. First, from a practical perspective if one can determine when repairs are more or less likely to be effective one can optimize the deployment of these repairs such that they occur at a time where they will have the greatest impact. This gives designers an important empowering them to design robots that can more predictably restore trust over time. Second, from a theoretical perspective answering this question could validate the original framework proposed in [21] and/or determine the different boundary conditions of this theoretical framework.

## 5.2 Study 2 - Combining Trust Repairs

In addition to repeated violations, the majority of previous research in this area has examined the simplest possible forms of apologies, denials, explanations, and promises [22]. Furthermore, no studies to date have fully compared the efficacy of combined trust repairs. This is problematic as the use of two distinct repair strategies in combination may complement, substitute or contradict with one another leading to unpredictable results. For example, if may be that by combining apologies with promises both perceptions of a robot's benevolence and ability increase equally but, it is equally possible that these two repairs cancel each other out. Furthermore it may also be that one strategy supersedes the other only leading to increases in ability *or* benevolence. This leads me to my second research question :

> **Study 2 | RQ:** Which combination of trust repair strategies are likely to be effective in restoring trust and how?

Answering this question is critical as combining trust repair strategies may ultimately lead to more effective trust repairs. Furthermore, by examining these repairs at a deeper level, these combined – and potentially multi-dimensional – effects may become more salient. This would contribute to the literature in two ways. First, the findings of this study would have practical application as they would allow designers to more effectively optimize their selection of robot trust repair strategies. Second, form a theoretical perspective, these findings can leverage the results of our first proposed study and add to our understanding not only of what strategies repair trust but if by combining these strategies the same mechanisms appear of if new or contradictory impacts emerge. This can be achieved through mapping each individual repair strategy to a specific sub-component of trustworthiness and seeing if when combined these mappings hold consistent or shift.

## 5.3 Study 3 - Ordering Trust Repairs

Finally, the majority of studies examining trust repairs over multiple violations have used the same strategies repeatedly [22]. For example, when examining the impact of apologies over time participants may be exposed to three subsequent apologies rather than an apology after the first violation, a promise after the second, and an explanation after the third. While this has allowed scholars to simplify study designs and more closely examine these strategies in isolation, this has prevented additional explorations into how changing the order of repair strategies may impact their efficacy. Indeed, it may be that some repairs are more effective when deployed after other repairs. For example, it is possible that an apology delivered after a promise is more effective than a promise delivered after an apology. Alternatively, the inverse may also be true and some repairs may be less effective after others. To date, however, no studies to my knowledge have considered how varying the order of repairs strategies over multiple violations of trust. This leads me to my third research question:

> **Study 3 | RQ:** When there are repeated trust violations, is the effectiveness of trust repairs related to the order they are given, if so what order is the most effective?

Answering this question will be vital in enabling designers and developers of robots to adapt their trust repair approaches based on previous responses. Alternatively, if previous repair has no impact on the efficacy of subsequent repairs, then designers may likewise be free to use a consistent repair strategy. In addition to these practical implications, by leveraging the results of the first proposed study in this series, it is also possible to examine if the same mechanisms remain consistent independent of ordering effects or if repairs shift in what element of trustworthiness they impact. This has important theoretical implications as it could produce a unique boundary condition for current theories (including that proposed in [21]) on the topic of trust repair in general and more specifically in HRI.

## 6 Conclusion

This extended abstract provides an overview of the fundamental concepts and key components of trust repair in human-robot interaction. It also summarizes some of my current published work in this field, offering a context for their future research plans, including associated research questions and potential contributions. Specifically, this upcoming work aims to investigate the mechanisms of trust repair and the impact of different combinations and orders of these repairs on overall efficacy. This research has the potential to enable designers to create more trustworthy robots by improving their ability to restore trust. Additionally, the findings may contribute to theoretical development and identify the circumstances in which certain frameworks are most applicable.

7A PREPRINT - JULY 14, 2023[18] Yaohui Guo and X. Jessie Yang. Modeling and predicting trust dynamics in human–robot teaming: A bayesian inference approach. *International Journal of Social Robotics*, 13, 2021.

[19] Roy J Lewicki and Chad Brinsfield. Trust repair. *Annual Review of Organizational Psychology and Organizational Behavior*, 4:287–313, 2017.

[20] Roderick M Kramer and Roy J Lewicki. Repairing and enhancing trust: Approaches to reducing organizational trust deficits. *Academy of Management annals*, 4(1):245–277, 2010.

[21] Connor Esterwood and Lionel P Robert Jr. Three strikes and you are out!: The impacts of multiple human–robot trust violations and repairs on robot trustworthiness. *Computers in Human Behavior*, 142:107658, 2023.

[22] Connor Esterwood and Lionel P. Robert. A literature review of trust repair in hri. In *Proceedings of 31th IEEE International Conference on Robot and Human Interactive Communication*, ROMAN '22. IEEE Press, 2022.

[23] Lionel P Robert Jr, Alan R Dennis, and Manju K Ahuja. Differences are different: Examining the effects of communication media on the impacts of racial and gender diversity in decision-making teams. *Information Systems Research*, 29(3):525–545, 2018.

[24] Lionel Robert and Sangseok You. Are you satisfied yet? shared leadership, trust and individual satisfaction in virtual teams. In *Proceedings of the iConference*, 2013.

[25] Wonjoon Kim, Nayoung Kim, Joseph B Lyons, and Chang S Nam. Factors affecting trust in high-vulnerability human-robot interaction contexts: A structural equation modelling approach. *Applied ergonomics*, 85:103056, 2020.

[26] A. Costa, D. Ferrin, and C. Fulmer. Trust at work. *The sage handbook of industrial, work & organizational psychology*, pages 435–467, 2018.

[27] Kurt T Dirks and Daniel P Skarlicki. The relationship between being perceived as trustworthy by coworkers and individual performance. *Journal of Management*, 35(1):136–157, 2009.

[28] Kinshuk Sharma, F. David Schoorman, and Gary A. Ballinger. How can it be made right again? a review of trust repair research. *Journal of Management*, 0(0):01492063221089897, 2022.

[29] Anthony L Baker, Elizabeth K Phillips, Daniel Ullman, and Joseph R Keebler. Toward an understanding of trust repair in human-robot interaction: current research and future directions. *ACM Transactions on Interactive Intelligent Systems (TiiS)*, 8(4):1–30, 2018.

[30] Akuadasuo Ezenyilimba, Margaret Wong, Alexander Hehr, Mustafa Demir, Alexandra Wolff, Erin Chiou, and Nancy Cooke. Impact of transparency and explanations on trust and situation awareness in human–robot teams. *Journal of Cognitive Engineering and Decision Making*, page 15553434221136358, 2022.

[31] Brad R Rawlins. Measuring the relationship between organizational transparency and employee trust. 2008.

[32] Maurice E Schweitzer, John C Hershey, and Eric T Bradlow. Promises and lies: Restoring violated trust. *Organizational behavior and human decision processes*, 101(1):1–19, 2006.

[33] Paul Robinette, Ayanna M Howard, and Alan R Wagner. Timing is key for robot trust repair. In *International conference on social robotics*, pages 574–583. Springer, 2015.

[34] E. S. Kox, J. H. Kerstholt, T. F. Hueting, and P. W. De Vries. Trust repair in human-agent teams: the effectiveness of explanations and expressing regret. *Autonomous Agents and Multi-Agent Systems*, 35(2), 2021.

[35] Connor Esterwood and Lionel P Robert. Having the right attitude: How attitude impacts trust repair in human-robot interaction. In *Proceedings of the 2022 ACM/IEEE International Conference on Human-Robot Interaction (HRI 2022)*. ACM/IEEE, 2022.

[36] Xinyi Zhang. "sorry, it was my fault": Repairing trust in human-robot interactions. Master's thesis, University of Oklahoma, 5 2021.

[37] Sarah Strohkorb Sebo, Priyanka Krishnamurthi, and Brian Scassellati. "i don't believe you": Investigating the effects of robot trust violation and repair. In *2019 14th ACM/IEEE International Conference on Human-Robot Interaction (HRI)*, pages 57–65. IEEE, 2019.

[38] Connor Esterwood and Lionel P Robert. Do you still trust me? human-robot trust repair strategies. In *2021 30th IEEE International Conference on Robot & Human Interactive Communication (RO-MAN)*, pages 183–188. IEEE, 2021.

[39] Filipa Correia, Carla Guerra, Samuel Mascarenhas, Francisco S Melo, and Ana Paiva. Exploring the impact of fault justification in human-robot trust. In *Proceedings of the 17th international conference on autonomous agents and multiagent systems*, pages 507–513, 2018.